\documentclass{article}
\usepackage{arxiv}

\usepackage[utf8]{inputenc} 
\usepackage[T1]{fontenc}    
\usepackage{hyperref}       
\usepackage{url}            
\usepackage{booktabs}       
\usepackage{amsfonts}       
\usepackage{nicefrac}       
\usepackage{microtype}      
\usepackage{lipsum}
\usepackage{float}
\usepackage{graphicx}
\usepackage{multirow}
\usepackage{array}
\usepackage{comment}
\usepackage{longtable}
\usepackage{ragged2e}
\usepackage{subfig}
\usepackage[table]{xcolor}
\usepackage{booktabs}
\usepackage{siunitx}
\usepackage{amsmath}
\usepackage{geometry}
\usepackage{multicol}
\usepackage{amsmath}
\usepackage{pifont}
\title{Physics-Informed Multimodal Bearing Fault Classification under Variable Operating Conditions using Transfer Learning}

\author{
    Tasfiq E. Alam \\
    Department of Industrial and Systems Engineering \\
    University of Oklahoma \\
    Norman, Oklahoma-73019 \\
    \texttt{tasfiq@ou.edu} \\
    \And
    Md Manjurul Ahsan \\
    Department of Industrial and Systems Engineering \\
    University of Oklahoma \\
    Norman, Oklahoma-73019 \\
    \texttt{ahsan@ou.edu} \\
    \And
    Shivakumar Raman \\
    Department of Industrial and Systems Engineering \\
    University of Oklahoma \\
    Norman, Oklahoma-73019 \\
    \texttt{raman@ou.edu} \\
}

\begin{document}
\maketitle

\begin{abstract}
Accurate and interpretable bearing fault classification is critical for ensuring the reliability of rotating machinery, particularly under variable operating conditions where domain shifts can significantly degrade model performance. This study proposes a physics-informed multimodal convolutional neural network (CNN) with a late fusion architecture, integrating vibration and motor current signals alongside a dedicated physics-based feature extraction branch. The model incorporates a novel physics-informed loss function that penalizes physically implausible predictions based on characteristic bearing fault frequencies---Ball Pass Frequency Outer (BPFO) and Ball Pass Frequency Inner (BPFI)---derived from bearing geometry and shaft speed. Comprehensive experiments on the Paderborn University dataset demonstrate that the proposed physics-informed approach consistently outperforms a non-physics-informed baseline, achieving higher accuracy, reduced false classifications, and improved robustness across multiple data splits. To address performance degradation under unseen operating conditions, three transfer learning (TL) strategies---Target-Specific Fine-Tuning (TSFT), Layer-Wise Adaptation Strategy (LAS), and Hybrid Feature Reuse (HFR)---are evaluated. Results show that LAS yields the best generalization, with additional performance gains when combined with physics-informed modeling. Validation on the KAIST bearing dataset confirms the framework’s cross-dataset applicability, achieving up to 98\% accuracy. Statistical hypothesis testing further verifies significant improvements ($p < 0.01$) in classification performance. The proposed framework demonstrates the potential of integrating domain knowledge with data-driven learning to achieve robust, interpretable, and generalizable fault diagnosis for real-world industrial applications.
\end{abstract}


\keywords{
Physics-informed learning \and
Bearing fault diagnosis \and
Convolutional neural network \and
Transfer learning \and
Condition monitoring \and
Domain adaptation
}

\section{Introduction}

Rotating machinery—found in infrastructures ranging from motors and turbines to automotive engines—is critical to industrial systems. Failures in such equipment can cause unplanned downtime, costly repairs, and safety risks. Effective fault diagnosis is thus essential to ensure operational reliability and support condition-based maintenance strategies \cite{inyang2023diagnosis,wang2022review}.  

\begin{figure}[H]
    \centering
    \includegraphics[width=0.45\textwidth]{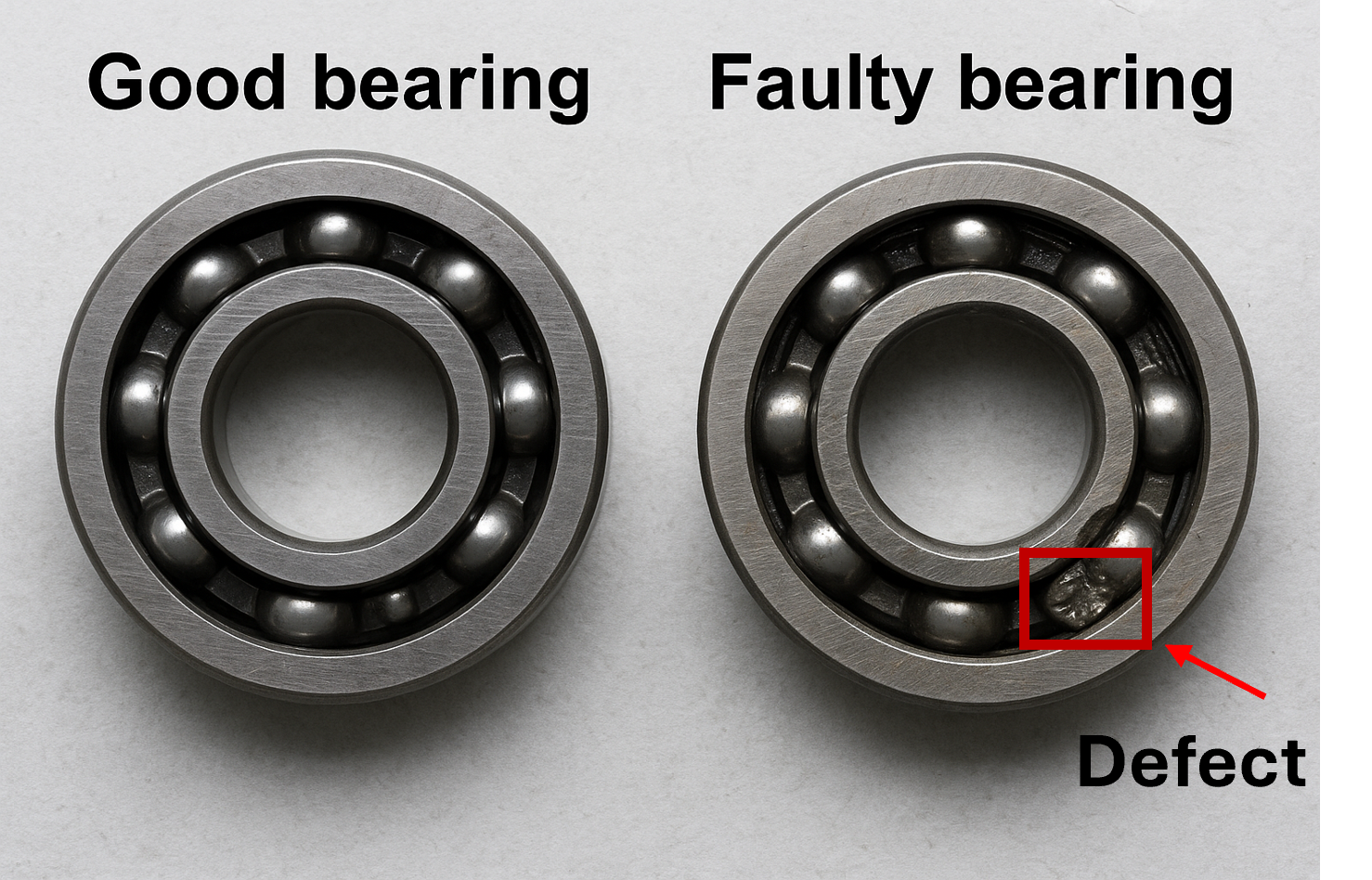}
    \caption{Comparison between a healthy bearing (left) and a faulty bearing (right) showing surface wear and spalling.}
    \label{fig:bearing_example}
\end{figure}

Figure~\ref{fig:bearing_example} illustrates a comparison between a healthy bearing and a faulty bearing, highlighting visible surface damage such as spalling and wear.

Traditional fault diagnosis approaches rely heavily on vibration-based condition monitoring techniques, such as spectral analysis, envelope analysis, and time–frequency methods, to detect abnormal signals characteristic of defects such as bearing faults or rotor imbalance \cite{mohd2021vibration}. With the advent of deep learning, data-driven models—particularly convolutional neural networks (CNNs)—have improved fault detection performance by automatically learning features from raw sensor data \cite{tama2023recent}. However, such purely data-driven models may struggle to generalize when operating conditions change, or when data scarcity, noise, or domain shifts occur \cite{liu2022survey}.  

Physics-informed machine learning (PIML) has emerged as a compelling alternative, offering the potential to fuse domain knowledge with data-driven approaches. By incorporating physical laws or constraints into model architectures or loss functions, PIML improves both interpretability and generalization—especially in condition monitoring tasks influenced by underlying physical mechanisms \cite{meng2025physics}.  

In rotating machinery, characteristic frequencies like BPFO (Ball Pass Frequency Outer) and BPFI (Ball Pass Frequency Inner) are rooted in bearing geometry and shaft rotation dynamics. Embedding these physical signatures into models can enhance diagnostic accuracy and reduce implausible misclassifications.

In this paper, we propose a physics-informed multimodal CNN that integrates a dedicated branch for physics-based feature extraction—focused on BPFO and BPFI—into a late fusion architecture combining vibration and motor current signals. To further enforce physical consistency, we introduce a novel physics-informed loss function penalizing predictions inconsistent with expected spectral signatures.

We also address domain shift through three transfer learning (TL) strategies—Target-Specific Fine-Tuning (TSFT), Layer-Wise Adaptation Strategy (LAS), and Hybrid Feature Reuse (HFR)—to enable robust generalization under varying operating conditions. Validation across the Paderborn and KAIST bearing datasets demonstrates significant performance gains, supported by statistical significance tests.

The main contributions of this work are:
\begin{itemize}
  \item A multimodal CNN with a physics-informed branch and loss formulation to improve interpretability and robustness in fault classification.
  \item Comparative analysis of three TL strategies for domain adaptation under variable operating conditions.
  \item Cross-dataset validation and statistical verification of performance improvements.
\end{itemize}

\section{Related Work}

Bearing fault diagnosis has traditionally relied on vibration-based condition monitoring methods such as spectral and envelope analysis to detect characteristic fault frequencies like BPFO and BPFI. These methods require expert feature engineering and can be sensitive to noise or changes in operating conditions. With the advent of deep learning, convolutional neural networks (CNNs) have gained popularity for their ability to automatically extract hierarchical features from raw sensor data, eliminating the need for manual feature design. Comprehensive reviews by Zhang et al.\,(2019) and Zhu (2023) illustrate the superiority of CNNs over traditional ML models in fault classification \cite{zhao2025comprehensive,zhang2019bearing}. However, purely data-driven models may struggle with domain shifts caused by variations in load, speed, or environmental conditions.

Physics-informed machine learning (PIML) has emerged as a promising strategy to enhance robustness and interpretability in fault diagnosis. By embedding physical knowledge—such as characteristic fault frequencies and threshold behavior—into model architecture or loss, PIML constrains predictions toward physically plausible outcomes. Shen et al.\,(2021) applied a CNN with a physics-informed threshold to bearing diagnosis, demonstrating improved reliability under variable conditions \cite{shen2021physics}. Hassannejad et al.\,(2025) extended such concepts with physics-guided wavelet-processing layers, reinforcing the importance of domain knowledge integration \cite{hassannejad2025adaptive}.

Simultaneously, transfer learning (TL) has been widely used to address data scarcity and domain shifts in condition monitoring. Tong et al.\,(2018) introduced a domain adaptation method using MMD to align feature distributions across operating conditions \cite{tong2018bearing}. Asutkar et al.\,(2023) demonstrated that applying TL with CNNs significantly boosts diagnostic robustness under variable environments \cite{asutkar2023deep}. Xu et al.\,(2025) proposed MCRCNet, a multi-scale CNN using TL, which improved generalization across different datasets \cite{xu2025mcrcnet}.

Despite these advances, most studies either focus on physics-informed modeling or domain adaptation, but rarely both. Physics-informed methods enhance explainability and constraints, yet often lack mechanisms to accommodate changing domains. TL methods improve generalization but may ignore physical consistency. This work bridges the gap by combining a physics-informed multimodal CNN—featuring BPFO/BPFI-based feature extraction and a tailored loss function—with three transfer learning strategies to achieve robust, interpretable, and generalizable fault classification under variable operating conditions and across datasets.
\section{Methodology}

\subsection{Overview of the Proposed Framework}
\begin{figure}[h!]
    \centering
    \includegraphics[width=0.95\textwidth]{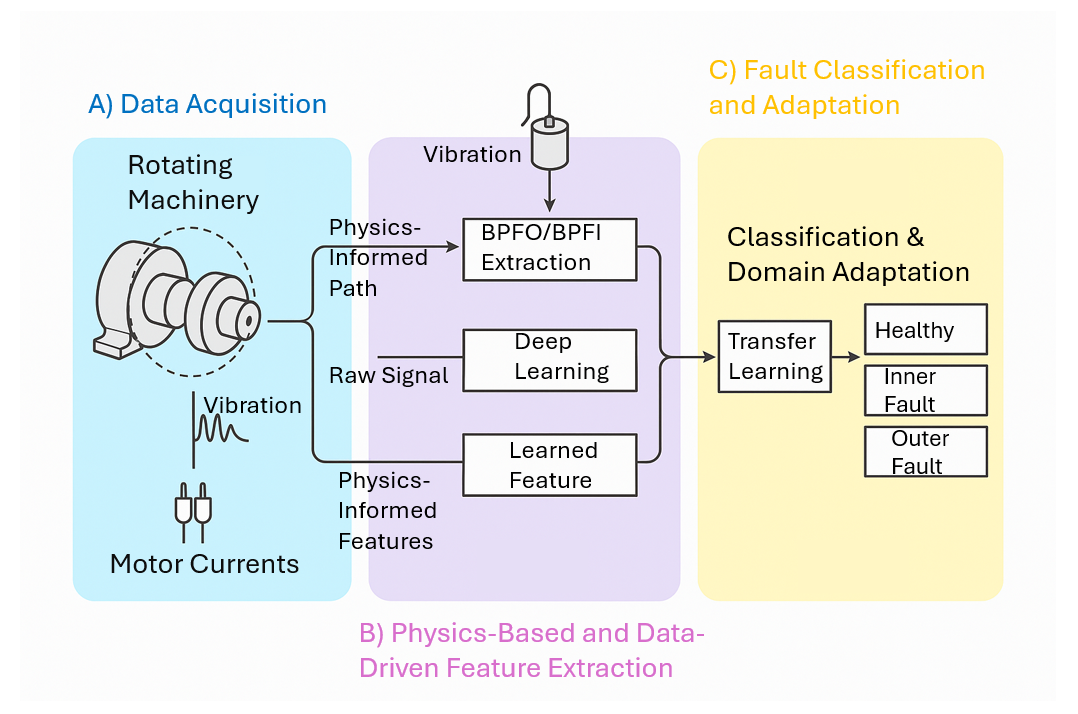} 
    \caption{
    Overview of the proposed physics-informed multimodal bearing fault classification framework. 
    \textbf{(A) Data Acquisition:} Vibration and motor current signals are collected from rotating machinery using dedicated sensors. 
    \textbf{(B) Physics-Based and Data-Driven Feature Extraction:} Physics-informed features are derived via BPFO/BPFI extraction and signal envelope analysis, while deep learning modules extract data-driven features from raw signals. 
    \textbf{(C) Classification and Adaptation:} A transfer learning module integrates the multimodal features to classify operating conditions into Healthy, Inner Fault, or Outer Fault, enabling adaptation to new domains.
    }
    \label{fig:workflow}
\end{figure}

The proposed framework integrates domain knowledge of bearing dynamics into a multimodal deep learning architecture to improve both interpretability and robustness in fault classification tasks. As illustrated in Figure~\ref{fig:workflow}, the proposed framework consists of three main stages: data acquisition, physics-based and data-driven feature extraction, and classification with domain adaptation.

\subsection{Data Acquisition}
This study employs the publicly available Paderborn University (PU) bearing dataset \cite{Lessmeier2016}, which contains both vibration and motor current signals collected under controlled laboratory conditions. The dataset comprises measurements from healthy bearings as well as bearings with artificially induced faults, including single-point damage to the inner race, outer race, and rolling elements.  

Each bearing is tested under multiple operating conditions, with shaft speeds ranging from 900 to 1500~rpm and radial loads between 0.7~kN and 1.4~kN. Vibration signals are recorded using high-resolution accelerometers mounted in the radial and axial directions, while motor current signals are simultaneously captured from the drive motor.  

For each faulty bearing, the geometric parameters are provided, enabling calculation of characteristic fault frequencies such as the Ball Pass Frequency Outer (BPFO), Ball Pass Frequency Inner (BPFI), Ball Spin Frequency (BSF), and Fundamental Train Frequency (FTF). These physics-based quantities form the foundation for the hybrid feature extraction process used in the proposed framework.

\subsection{Physics-Based Feature Extraction}
Rolling element bearings generate two main types of vibration during operation: (1) high-frequency inherent vibrations caused by the elastic properties of the bearing, which are always present; and (2) defect-induced impact vibrations, which occur only after damage develops in the bearing components. Defects in the outer or inner race produce characteristic frequencies in the vibration spectrum, known as BPFO and BPFI, respectively. These are defined as \cite{harris2021rolling}:

\[
\text{BPFO} = \frac{n}{2} f_r \left(1 - \frac{d}{D} \cos \phi \right)
\]
\[
\text{BPFI} = \frac{n}{2} f_r \left(1 + \frac{d}{D} \cos \phi \right)
\]

where:
\begin{itemize}
    \item $n$ = number of rolling elements (8 in this study),
    \item $d$ = diameter of rolling element (6.75~mm),
    \item $D$ = pitch diameter of the bearing (28.55~mm),
    \item $f_r$ = shaft rotational frequency (Hz),
    \item $\phi$ = contact angle between load direction and rolling element centerline.
\end{itemize}

\begin{figure}[htbp]
    \centering
    \includegraphics[width=0.85\textwidth]{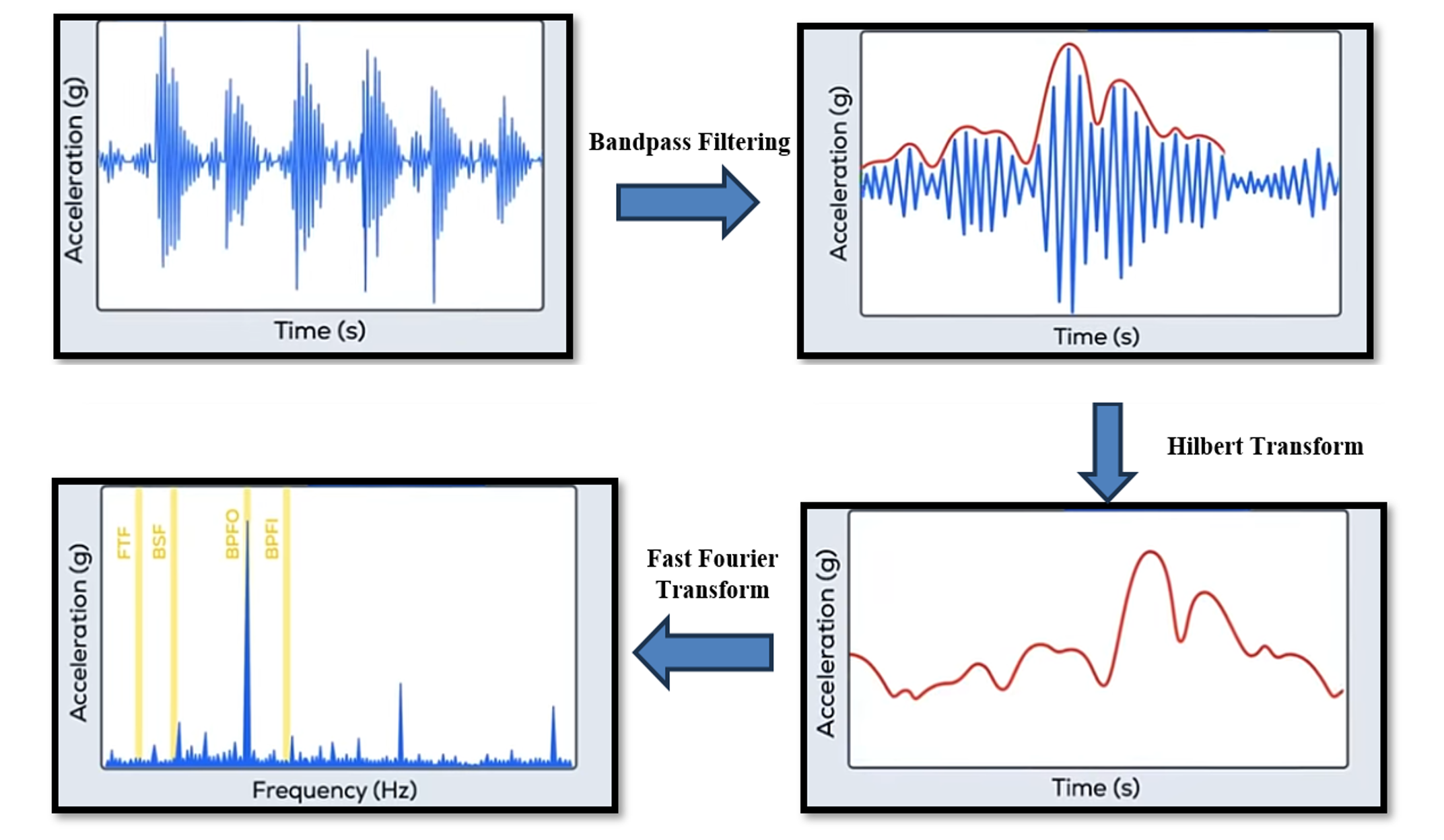}
    \caption{Envelope spectrum analysis procedure for extracting physically meaningful fault frequencies (adapted from \cite{randall2011rolling}).}
    \label{fig:envelope}
\end{figure}

To isolate these characteristic frequencies, envelope spectrum analysis is applied to the vibration signal \cite{randall2011rolling}. As illustrated in Figure~\ref{fig:envelope}, this process includes:
\begin{enumerate}
    \item Bandpass filtering to remove noise and low-frequency trends.
    \item Hilbert transform to obtain the signal envelope.
    \item Fast Fourier Transform (FFT) of the envelope to reveal spectral peaks at BPFO and BPFI.
\end{enumerate}
These extracted amplitudes are later incorporated into the physics-informed loss function.

\subsection{Physics-Informed Loss Function}
While conventional CNN models are optimized using only data-driven loss functions such as cross-entropy, our approach incorporates an additional penalty term to enforce physical consistency. Inspired by the physics-informed learning framework of Karniadakis et al.~(2021) \cite{karniadakis2021physics}, the penalty is applied when the model predicts an ``Outer Fault'' or ``Inner Fault'' but the envelope spectrum amplitude at BPFO or BPFI, respectively, falls below a defined threshold.

Let $A_{\text{BPFO},i}$ and $A_{\text{BPFI},i}$ be the envelope amplitudes at BPFO and BPFI for sample $i$, and $T_{\text{BPFO}}$, $T_{\text{BPFI}}$ be the corresponding thresholds. The penalty term $P_i$ is defined as:
\[
P_i =
\begin{cases}
T_{\text{BPFO}} - A_{\text{BPFO},i}, & \text{if predicted class = Outer Fault and } A_{\text{BPFO},i} < T_{\text{BPFO}},\\
T_{\text{BPFI}} - A_{\text{BPFI},i}, & \text{if predicted class = Inner Fault and } A_{\text{BPFI},i} < T_{\text{BPFI}},\\
0, & \text{otherwise}.
\end{cases}
\]

The final loss for a batch is computed as:
\[
\mathcal{L} = \mathcal{L}_{\text{CE}} + \lambda \frac{1}{N} \sum_{i=1}^N P_i
\]
where $\mathcal{L}_{\text{CE}}$ is the sparse categorical cross-entropy loss, $\lambda$ is the penalty weight, and $N$ is the batch size.

\subsection{Multimodal CNN Architecture}
\begin{figure}[htbp]
    \centering
    \includegraphics[width=0.9\textwidth]{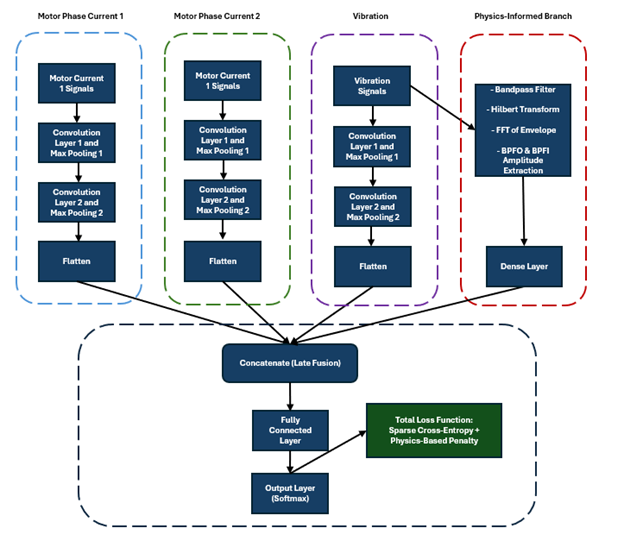}
    \caption{Proposed physics-informed multimodal CNN architecture with late fusion.}
    \label{fig:architecture}
\end{figure}

The proposed model extends a late fusion multimodal CNN architecture with a dedicated physics-informed branch. As shown in Figure~\ref{fig:architecture}, three parallel convolutional pipelines process:
\begin{enumerate}
    \item Motor current signal 1,
    \item Motor current signal 2,
    \item Vibration signal.
\end{enumerate}
In addition, the physics-informed branch processes vibration data through bandpass filtering, Hilbert transform, and FFT to extract normalized BPFO and BPFI amplitudes. These features pass through a dense layer before being concatenated with outputs from the other branches. The fused feature vector is processed through fully connected layers, followed by a softmax output layer. Training is guided by the combined data-driven and physics-informed loss function.

\subsection{Transfer Learning Strategies}

Although the baseline and physics-informed models in previous chapters achieved high accuracy under fixed baseline operating conditions, their performance degraded when applied to unseen variable operating conditions due to domain shifts in signal characteristics~\cite{yang2023domain,pan2009survey}. To address this challenge, three transfer learning (TL) strategies were investigated, adapting models trained on a baseline operating condition to additional operating conditions with varying radial loads, torque, and speeds. These strategies were applied to both non-physics-informed and physics-informed models. In addition to quantitative classification performance, t-SNE visualizations were employed to qualitatively assess feature distribution shifts before and after TL. Furthermore, the KAIST bearing dataset~\cite{lee2020kaist} was used to validate the TL framework on an independent dataset with different signal characteristics.

\subsubsection{Target-Specific Fine-Tuning (TSFT)}

In TSFT (Model~1), the pre-trained model (trained on the baseline condition \texttt{N15\_M07\_F10} of the Paderborn dataset~\cite{lessmeier2016condition}) was used for weight initialization. All convolutional layers were frozen, and only the fully connected layer was fine-tuned on target domain data. This approach preserves domain-invariant features learned from the source domain while adapting the classifier to the target condition~\cite{liu2021domain,wen2020transfer}.

The model parameters can be expressed as:
\[
\theta = (\theta_f, \theta_c)
\]
where $\theta_f$ represents frozen convolutional feature extractor layers, and $\theta_c$ denotes trainable classification layers.

The optimization objective is:
\[
\theta_c^* = \arg\min_{\theta_c} \frac{1}{N} \sum_{i=1}^{N} L\big(f(x_i; \theta_f, \theta_c), y_i\big)
\]
where:
\begin{itemize}
    \item $N$: number of training samples in the target domain
    \item $L$: loss function (e.g., Cross-Entropy Loss)
    \item $f(x_i; \theta_f, \theta_c)$: model output
    \item $(x_i, y_i)$: input sample and its true label
\end{itemize}

\begin{figure}[h!]
    \centering
    \includegraphics[width=0.8\textwidth]{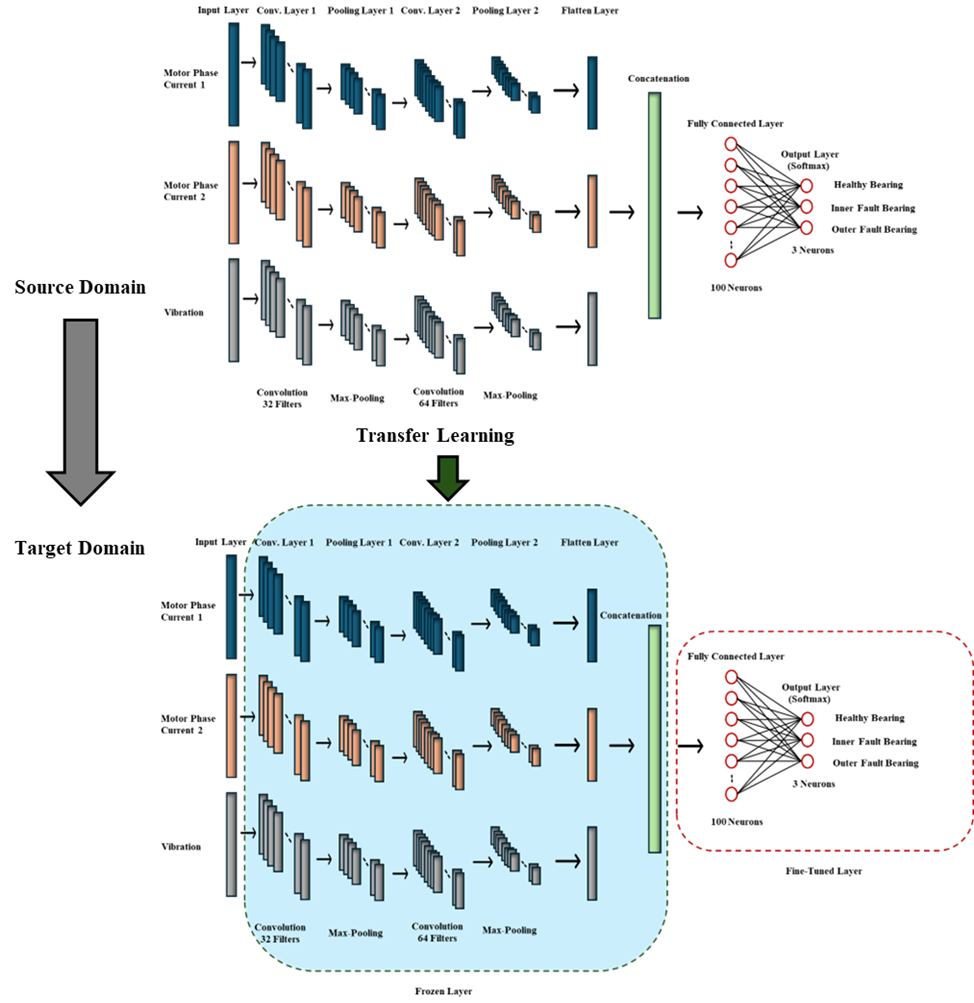}
    \caption{Framework of the proposed TL Model~1 (Target-Specific Fine-Tuning).}
    \label{fig:tsft}
\end{figure}

This configuration allows the model to retain robust, pre-learned features while adjusting only the decision boundary for the new operating condition, aligning with findings in prior TL studies~\cite{liu2021domain,pan2009survey}. The framework is illustrated in Figure~\ref{fig:tsft}.

\subsubsection{Other TL Strategies}

Two additional TL strategies were implemented and are presented in Appendix~\ref{appendix:tl_strategies}:

\begin{itemize}
    \item \textbf{Layer-Wise Adaptation Strategy (LAS):} freezes early convolution and pooling layers while fine-tuning deeper layers and the classifier~\cite{liu2021domain,wen2020transfer} (see Figure~\ref{fig:las} in Appendix~\ref{appendix:tl_strategies}).
    \item \textbf{Hybrid Feature Reuse (HFR):} freezes all layers before the fully connected stage and replaces the final classifier with a new target-specific layer~\cite{wen2020transfer} (see Figure~\ref{fig:hfr} in Appendix~\ref{appendix:tl_strategies}).
    \end{itemize}
\subsection{Experimental Setup}

Experiments are conducted using the Paderborn University bearing dataset~\cite{lessmeier2016condition} for model development and the KAIST bearing dataset~\cite{han2020bearing} for cross-dataset validation. 

From the Paderborn dataset, vibration and two-phase motor current signals were segmented using a sliding window of 10,000 samples with a stride of 5,000 samples. Each segment was standardized to zero mean and unit variance, resulting in a total of 23,825 segments:
\begin{itemize}
    \item 4,929 healthy samples
    \item 9,037 inner race fault samples
    \item 9,858 outer race fault samples
\end{itemize}

Unlike the baseline physics-informed experiments, which employed three train–test split ratios (80/20, 70/30, and 60/40), the transfer learning experiments used a fixed 60\% training, 20\% validation, and 20\% testing split to allow consistent adaptation and evaluation across models. For each split, the training portion was further divided into 85\% for model training and 15\% for validation during hyperparameter tuning.

\begin{table}[h!]
\centering
\caption{Dataset composition for different train–test splits of the Paderborn dataset.}
\label{tab:dataset_splits}
\begin{tabular}{lcccccc}
\hline
\multirow{2}{*}{\textbf{Class}} & 
\multicolumn{2}{c}{\textbf{80/20 Split}} & 
\multicolumn{2}{c}{\textbf{70/30 Split}} & 
\multicolumn{2}{c}{\textbf{60/40 Split}} \\
\cline{2-7}
& \textbf{Train} & \textbf{Test} & \textbf{Train} & \textbf{Test} & \textbf{Train} & \textbf{Test} \\
\hline
Healthy & 3,943 & 986 & 3,450 & 1,479 & 2,957 & 1,972 \\
Inner Race Fault & 7,230 & 1,807 & 6,326 & 2,711 & 5,422 & 3,615 \\
Outer Race Fault & 7,886 & 1,972 & 6,901 & 2,957 & 5,915 & 3,943 \\
\hline
\textbf{Total} & 19,059 & 4,765 & 16,677 & 7,147 & 14,294 & 9,530 \\
\hline
\end{tabular}
\end{table}

For all splits, the class distribution is kept balanced across the training and test sets to avoid introducing bias into the classification task. The same preprocessing procedure is also applied to the KAIST dataset, which is used for cross-domain evaluation under variable operating conditions.

\subsubsection{Model Training}
All models are trained using the Adam optimizer with a learning rate of $1\times 10^{-4}$, batch size of 32, and early stopping with a patience of 10 epochs. A maximum of 100 epochs is allowed. We use categorical cross-entropy loss for classification. Model weights are initialized using Xavier initialization. For the transfer learning experiments, the pre-trained model from the Paderborn dataset served as the source, with target-specific fine-tuning or other adaptation strategies applied in this study.

Performance is evaluated using Accuracy, Precision, Recall, F1-score, and the Area Under the ROC Curve (AUC), defined as~\cite{ahsan2020deep}:
\[
\text{Accuracy} = \frac{TP + TN}{TP + TN + FP + FN}
\]
\[
\text{Precision} = \frac{TP}{TP + FP}
\]
\[
\text{Recall} = \frac{TP}{TP + FN}
\]
\[
\text{F1-score} = 2 \times \frac{\text{Precision} \times \text{Recall}}{\text{Precision} + \text{Recall}}
\]
\[
\text{AUC} = \int_{0}^{1} TPR(FPR) \, d(FPR)
\]
where $TP$, $TN$, $FP$, and $FN$ denote true positives, true negatives, false positives, and false negatives, respectively.
\subsection{Hyperparameter Optimization}
A grid search was conducted to evaluate the influence of two hyperparameters in the physics-informed loss function: the penalty weight ($\lambda$) and the amplitude threshold at BPFO/BPFI for fault classification. The penalty weight adjusts the relative contribution of the physics term, while the threshold filters low-amplitude cases to reduce false positives.

The search space included:
\begin{itemize}
    \item Penalty ($\lambda$): \{0.05, 0.20, 1.00\}
    \item Threshold: \{5, 10, 15\} (percentiles of the amplitude distribution)
\end{itemize}

Each $(\lambda, \text{threshold})$ combination was trained and validated using 5-fold cross-validation under three train/test splits: 80/20, 70/30, and 60/40. The best fold for each split was selected based on validation accuracy, and final performance was reported on the held-out test set in terms of accuracy, training time, and per-class AUC. Results are shown in Table~\ref{tab:grid_search_results}.
\section{Results}
\subsection{PINN results}
A detailed grid search was performed to optimize the performance of the physics-informed model, focusing on two critical hyperparameters: the penalty weight ($\lambda$) and the amplitude threshold for the physics-informed loss term. The penalty weight regulates the impact of the physics-informed penalty within the overall loss function, whereas the threshold specifies the minimum amplitude necessary at BPFO/BPFI for the model to reliably assign fault labels. The grid search assessed the following values: Penalty ($\lambda$): [0.05, 0.2, 1.0] and Threshold: [5, 10, 15] (pertaining to certain percentiles of the amplitude distribution). Models were trained and validated for each combination using 5-fold cross-validation followed by three distinct train/test splits: 80/20, 70/30, and 60/40 using the Adam optimizer as described in Chapter~3. For each split, the optimal fold was chosen according to validation accuracy, and performance was assessed using test accuracy, training duration, and area under the ROC curve (AUC) for each class.

A thorough grid search was performed to evaluate the impact of the penalty coefficient ($\lambda$) and threshold values on the test accuracy of the physics-informed model across three different train-test splits: 80/20, 70/30, and 60/40. In general, lower threshold values (particularly a threshold of 5) were linked to improved test accuracy across all splits. For example, the 70/30 split resulted in the greatest test accuracy of 0.969 at $\lambda = 0.05$ and a threshold of 5. Similarly, in the 80/20 split, the best accuracy (0.958) was obtained with the identical parameter combination. For the 60/40 split, both $\lambda = 0.05$ and $\lambda = 0.20$ with a threshold of 5 achieved high accuracies of 0.956 and 0.958, respectively. 

\begin{table}[h!]
\centering
\caption{Impact of Penalty ($\lambda$) and Threshold on Model Accuracy for Different Data Splits}
\label{tab:grid_search_results}
\begin{tabular}{cccc}
\hline
\textbf{Split} & \textbf{Penalty ($\lambda$)} & \textbf{Threshold} & \textbf{Test Accuracy} \\
\hline
80/20 & 0.05 & 5  & 0.958 \\
      & 0.05 & 10 & 0.924 \\
      & 0.05 & 15 & 0.913 \\
      & 0.20 & 5  & 0.932 \\
      & 0.20 & 10 & 0.947 \\
      & 0.20 & 15 & 0.945 \\
      & 1.00 & 5  & 0.954 \\
      & 1.00 & 10 & 0.920 \\
      & 1.00 & 15 & 0.928 \\
\hline
70/30 & 0.05 & 5  & 0.969 \\
      & 0.05 & 10 & 0.939 \\
      & 0.05 & 15 & 0.911 \\
      & 0.20 & 5  & 0.939 \\
      & 0.20 & 10 & 0.942 \\
      & 0.20 & 15 & 0.945 \\
      & 1.00 & 5  & 0.933 \\
      & 1.00 & 10 & 0.963 \\
      & 1.00 & 15 & 0.956 \\
\hline
60/40 & 0.05 & 5  & 0.956 \\
      & 0.05 & 10 & 0.936 \\
      & 0.05 & 15 & 0.913 \\
      & 0.20 & 5  & 0.958 \\
      & 0.20 & 10 & 0.901 \\
      & 0.20 & 15 & 0.920 \\
      & 1.00 & 5  & 0.933 \\
      & 1.00 & 10 & 0.921 \\
      & 1.00 & 15 & 0.965 \\
\hline
\end{tabular}
\end{table}

Based on these factors, the configuration of $\lambda = 1.00$ and the 10th percentile threshold, as defined by the 70/30 split, was chosen for further analysis. At this setting, the test accuracy (0.963) is marginally lower than the highest accuracy (0.969) but remains among the top-performing configurations. Moreover, this configuration enforces a more robust physics-informed regularization on the model, guiding it toward physically plausible solutions while maintaining competitive predictive accuracy.

The performance of the physics-informed 1D CNN model with late fusion on the baseline operating condition is summarized in Table~\ref{tab:pi_model_perf}. The model consistently achieved accuracy, precision, recall, and F1-score of 0.95 across most test sizes and epochs, with only a slight drop to 0.93 for the 40\% test set at 10 epochs.

\begin{table}[h!]
\centering
\caption{Performance of the Physics-Informed Model on Baseline Operating Condition N15\_M07\_F10}
\label{tab:pi_model_perf}
\begin{tabular}{cccccc}
\hline
\textbf{Test Data} & \textbf{Epochs} & \textbf{Accuracy} & \textbf{Precision} & \textbf{Recall} & \textbf{F1-Score} \\
\hline
20\% & 10 & 0.95 & 0.95 & 0.95 & 0.95 \\
     & 50 & 0.95 & 0.95 & 0.95 & 0.95 \\
30\% & 10 & 0.95 & 0.95 & 0.95 & 0.95 \\
     & 50 & 0.95 & 0.95 & 0.95 & 0.95 \\
40\% & 10 & 0.93 & 0.93 & 0.93 & 0.93 \\
     & 50 & 0.95 & 0.95 & 0.95 & 0.95 \\
\hline
\end{tabular}
\end{table}

Confusion matrix analysis further confirmed the improved classification capability of the physics-informed model over the baseline. For the Healthy class, false positives decreased from 119 to 78, and false negatives dropped from 106 to 60. The Inner Fault class showed a slight increase in false positives (108 to 122) but a notable decrease in false negatives (192 to 166). For the Outer Fault class, false positives reduced from 173 to 122 and false negatives from 102 to 96, indicating enhanced fault-type discrimination and overall robustness.

\begin{figure}[h!]
    \centering
    \includegraphics[width=0.8\textwidth]{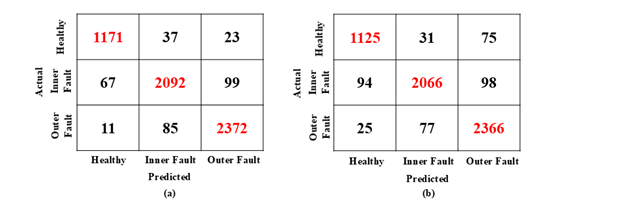}
    \caption{Confusion matrix of 80/20 split: (a) with Physics-Informed model and (b) without Physics-Informed model.}
    \label{fig:confusion_matrix_80_20}
\end{figure}

\subsection{Transfer Learning Results}

Table~\ref{tab:baseline_vs_conditions} presents the overall performance of the three operating conditions, which are used individually as the test set, while the baseline operating condition is used as the training set for each scenario. The table depicts that the variable operating conditions deteriorated the performance of the proposed CNN model on each of the metrics. The model performed best on the second operating condition with an accuracy of $0.90 \pm 0.01$, where rotational speed and radial force are consistent with the baseline condition, with the only alteration in load torque value from $0.7$~Nm to $0.1$~Nm. Conversely, condition 1 performs the worst with an accuracy of $0.40 \pm 0.01$, where parameters load torque and radial force are constant with the baseline condition, with the only change in rotational speed from 1500~rpm to 900~rpm. These performances from the variable operating conditions as the test set, with the pretrained model of the baseline operating condition as the train set, resulted in poor performance overall, barring the second operating condition.

\begin{table}[h!]
\centering
\caption{Computational results of using three operating conditions as an unseen test set with the baseline condition as the train set.}
\label{tab:baseline_vs_conditions}
\begin{tabular}{lcccc}
\hline
\textbf{Operating Condition} & \textbf{Accuracy} & \textbf{Precision} & \textbf{Recall} & \textbf{F1 Score} \\
\hline
N09\_M07\_F10 (1) & $0.40 \pm 0.01$ & $0.41 \pm 0.01$ & $0.49 \pm 0.01$ & $0.39 \pm 0.01$ \\
N15\_M01\_F10 (2) & $0.90 \pm 0.01$ & $0.89 \pm 0.01$ & $0.90 \pm 0.01$ & $0.89 \pm 0.01$ \\
N15\_M07\_F04 (3) & $0.76 \pm 0.01$ & $0.78 \pm 0.01$ & $0.76 \pm 0.01$ & $0.75 \pm 0.01$ \\
\hline
\end{tabular}
\end{table}

In Table~\ref{tab:cond1}, we have displayed the performance of the three TL-based models on operating condition 1. The table indicates that there is an improvement in performance for all the metrics in every TL model compared to the results presented in Table~\ref{tab:baseline_vs_conditions}. Model~2 showcased the best performance with accuracy of $0.92 \pm 0.01$, precision of $0.92 \pm 0.01$, recall of $0.92 \pm 0.01$, and F1 score of $0.92 \pm 0.01$ on the test set. Model~1 and Model~3 delivered similar results of approximately $0.81 \pm 0.01$ across all metrics. 

\begin{table}[h!]
\centering
\caption{The performance of three TL Models with a confidence interval $\alpha = 0.05$ on N09\_M07\_F10 (Operating Condition 1).}
\label{tab:cond1}
\begin{tabular}{lcccc}
\hline
\textbf{Model} & \textbf{Accuracy} & \textbf{Precision} & \textbf{Recall} & \textbf{F1 Score} \\
\hline
TSFT (Model 1) & $0.81 \pm 0.01$ & $0.81 \pm 0.01$ & $0.81 \pm 0.01$ & $0.81 \pm 0.01$ \\
LAS (Model 2) & $0.92 \pm 0.01$ & $0.92 \pm 0.01$ & $0.92 \pm 0.01$ & $0.92 \pm 0.01$ \\
HFR (Model 3) & $0.81 \pm 0.01$ & $0.82 \pm 0.01$ & $0.81 \pm 0.01$ & $0.81 \pm 0.01$ \\
\hline
\end{tabular}
\end{table}

Similarly, Table~\ref{tab:cond2} illustrates the performance of the TL-based models on the 2nd operating condition. Both Model~1 and Model~2 showed superior performance to the performance presented in Table~\ref{tab:baseline_vs_conditions}. Model~2 exhibited the best performance with a value of $0.96 \pm 0.01$ across all the metrics on the test data. In contrast, Model~3 performed worse than the results displayed in Table~\ref{tab:baseline_vs_conditions}, with an accuracy of $0.81 \pm 0.01$, precision of $0.82 \pm 0.01$, recall of $0.81 \pm 0.01$, and F1 Score of $0.81 \pm 0.01$. 

\begin{table}[h!]
\centering
\caption{The performance of three TL Models with a confidence interval $\alpha = 0.05$ on N15\_M01\_F10 (Operating Condition 2).}
\label{tab:cond2}
\begin{tabular}{lcccc}
\hline
\textbf{Model} & \textbf{Accuracy} & \textbf{Precision} & \textbf{Recall} & \textbf{F1 Score} \\
\hline
TSFT (Model 1) & $0.92 \pm 0.01$ & $0.92 \pm 0.01$ & $0.92 \pm 0.01$ & $0.92 \pm 0.01$ \\
LAS (Model 2) & $0.96 \pm 0.01$ & $0.96 \pm 0.01$ & $0.96 \pm 0.01$ & $0.96 \pm 0.01$ \\
HFR (Model 3) & $0.81 \pm 0.01$ & $0.82 \pm 0.01$ & $0.81 \pm 0.01$ & $0.81 \pm 0.01$ \\
\hline
\end{tabular}
\end{table}

Likewise, Table~\ref{tab:cond3} details the performance of the TL-based model on operating condition 3. Model~2 emerged as the top performer with a value of $0.89 \pm 0.01$ across all metrics on the test data, whereas Models~1 and~3 displayed the same performance with an accuracy of $0.83 \pm 0.01$, precision of $0.84 \pm 0.01$, recall of $0.83 \pm 0.01$, and F1 Score of $0.83 \pm 0.01$. 

\begin{table}[h!]
\centering
\caption{The performance of three TL Models with a confidence interval $\alpha = 0.05$ on N15\_M07\_F04 (Operating Condition 3).}
\label{tab:cond3}
\begin{tabular}{lcccc}
\hline
\textbf{Model} & \textbf{Accuracy} & \textbf{Precision} & \textbf{Recall} & \textbf{F1 Score} \\
\hline
TSFT (Model 1) & $0.83 \pm 0.01$ & $0.84 \pm 0.01$ & $0.83 \pm 0.01$ & $0.83 \pm 0.01$ \\
LAS (Model 2) & $0.89 \pm 0.01$ & $0.89 \pm 0.01$ & $0.89 \pm 0.01$ & $0.89 \pm 0.01$ \\
HFR (Model 3) & $0.83 \pm 0.01$ & $0.84 \pm 0.01$ & $0.83 \pm 0.01$ & $0.83 \pm 0.01$ \\
\hline
\end{tabular}
\end{table}

In contrast, Model~1 and Model~3 show relatively lower and more uncertain performance, as reflected by wider confidence intervals. These statistical insights provide evidence for the selection of Model~2 as the most reliable strategy for transfer learning in this setting.

Table~\ref{tab:train_time} presents the summary of the total trainable parameters and total process time to train the models that were employed on the baseline model as well as the transfer learning models. For comparison, all the data used in the models were given an equal split, with $60\%$ assigned for train set, $20\%$ for validation set, and $20\%$ for test set across the conditions. The Model~3 in all the conditions required the least time to train, with a time of $183.03$~s for condition~1, $182.94$~s for condition~2, and $184.13$~s for condition~3. Model~2 took the most time across operating conditions since it needed to train more parameters. The processing times are $830.15$~s, $835.64$~s, and $850.44$~s, respectively, for operating conditions 1 to 3. The Baseline Model showed a processing time of $848.73$~s. These findings indicate the variability in processing time across different models and emphasize the effectiveness of the TL-based models for efficient training.

\begin{table}[h!]
\centering
\caption{Total trainable parameters and computational time for the training phase of the baseline model and TL models.}
\label{tab:train_time}
\begin{tabular}{lccc}
\hline
\textbf{Condition} & \textbf{Model} & \textbf{Trainable Parameters} & \textbf{Process Time (s)} \\
\hline
Baseline Condition & Baseline Model & 47,981,011 & 848.73 \\
 & Model 1 & 47,962,003 & 185.09 \\
Condition 1 & Model 2 & 47,981,011 & 830.15 \\
 & Model 3 & 47,962,003 & 183.03 \\
Condition 2 & Model 1 & 47,962,003 & 184.01 \\
 & Model 2 & 47,981,011 & 835.64 \\
 & Model 3 & 47,962,003 & 182.94 \\
Condition 3 & Model 1 & 47,962,003 & 185.59 \\
 & Model 2 & 47,981,011 & 850.44 \\
 & Model 3 & 47,962,003 & 184.13 \\
\hline
\end{tabular}
\end{table}

\begin{figure}[h!]
    \centering
    \includegraphics[width=0.85\textwidth]{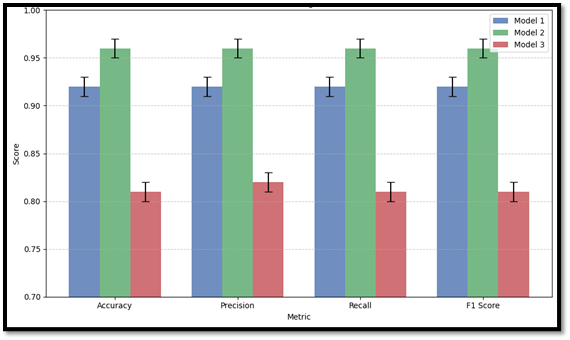}
    \caption{The performance of three TL Models with a CI $\alpha = 0.05$ on Operating Condition 1.}
    \label{fig:cond1_perf}
\end{figure}

\begin{figure}[h!]
    \centering
    \includegraphics[width=0.85\textwidth]{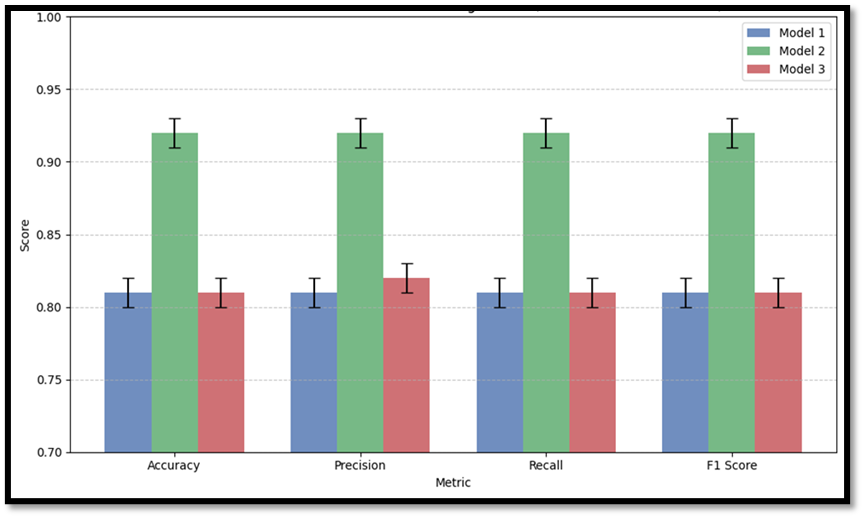}
    \caption{The performance of three TL Models with a CI $\alpha = 0.05$ on Operating Condition 2.}
    \label{fig:cond2_perf}
\end{figure}

\begin{figure}[h!]
    \centering
    \includegraphics[width=0.85\textwidth]{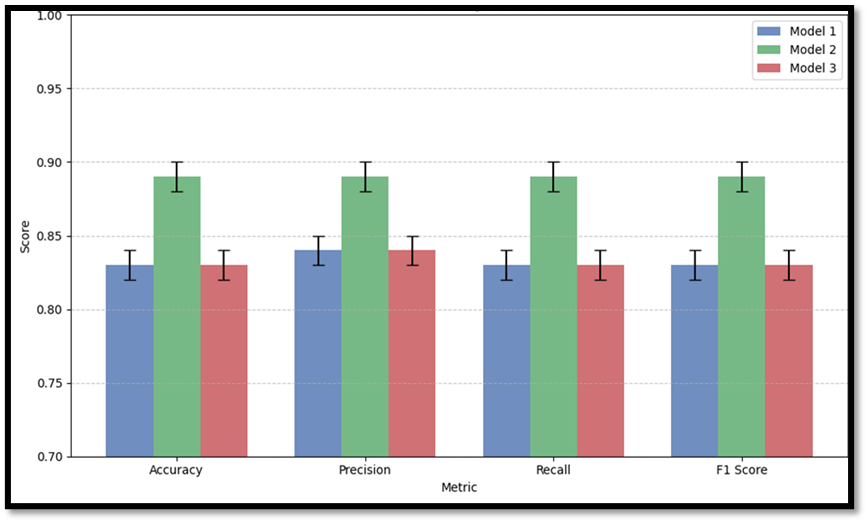}
    \caption{The performance of three TL Models with a CI $\alpha = 0.05$ on Operating Condition 3.}
    \label{fig:cond3_perf}
\end{figure}
\subsection{Feature visualization using t-SNE}
t-Distributed Stochastic Neighbor Embedding (t-SNE) is a widely utilized nonlinear dimensionality reduction method, especially effective for visualizing high-dimensional data in lower dimensions, such as two or three dimensions (van der Maaten \& Hinton, 2008). t-SNE is extensively applied in domains such as bioinformatics, image analysis, and fault diagnostics because of its robust visualization capabilities (Wattenberg et al., 2016).
To further understand the internal feature representations learned by our proposed model, we utilized t-SNE to visualize the high-dimensional fused feature vectors in a two-dimensional space.

\begin{figure}[h!]
    \centering
    \includegraphics[width=0.75\textwidth]{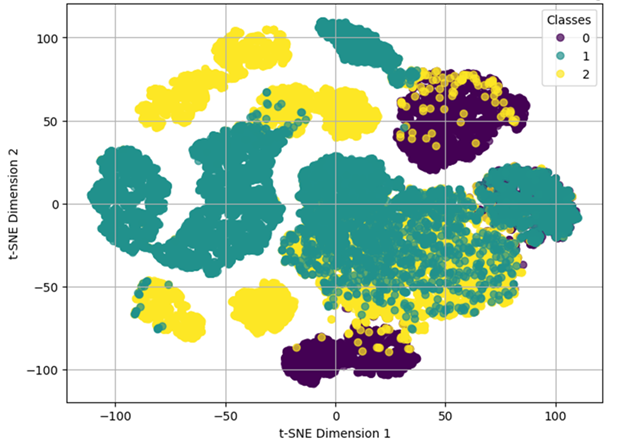}
    \caption{t-SNE visualization of feature distributions for the Baseline Model after training under the baseline condition.}
    \label{fig:tsne_baseline}
\end{figure}

\begin{figure}[h!]
    \centering
    \includegraphics[width=0.75\textwidth]{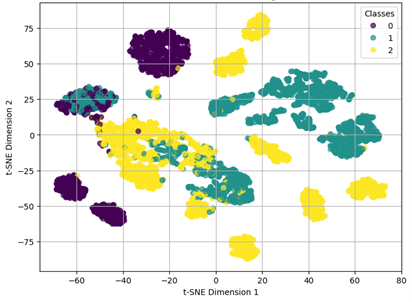}
    \caption{t-SNE visualization of feature distributions for Transfer Learning Model~2 after adaptation to a new operating condition.}
    \label{fig:tsne_tl}
\end{figure}

Figure~\ref{fig:tsne_baseline} and Figure~\ref{fig:tsne_tl} present the t-SNE visualization of feature distributions for the Baseline Model after training under the baseline condition while testing on an unseen test set (different operating condition) and the Transfer Learning Model after adaptation to a new operating condition. The plots demonstrate clear clustering patterns corresponding to Healthy, Inner Fault, and Outer Fault categories, validating that the model learned discriminative features. After transfer learning, the feature clusters exhibit tighter grouping and better separation, confirming effective knowledge transfer across different operating conditions.

\subsection{Validation on KAIST dataset}
To further validate the generalization capability of the proposed multimodal 1D CNN architecture beyond the Paderborn University dataset, we conducted additional experiments on the KAIST bearing dataset. The KAIST bearing dataset, developed by the Korea Advanced Institute of Science and Technology (KAIST), provides multimodal sensor data for bearing fault diagnosis under realistic and variable operating conditions. The dataset comprises synchronized vibration and motor current signals (similar to the modalities used in the Paderborn dataset) collected from a test rig equipped with deep groove ball bearings subjected to varying rotational speeds and load levels. It is extensively utilized for machine learning and deep learning applications in bearing fault classification.

The Model~1 transfer learning strategy, which freezes all convolutional and pooling layers and fine-tunes the fully connected layer, was applied for the KAIST experiments at the setting of rotational speed 3010~rpm, 0 load torque, and 0 radial force. The preprocessing steps, including modality-specific standardization and sliding window segmentation (window size = 10{,}000; stride = 5{,}000), were identical to those used for the Paderborn data to ensure consistency. After fine-tuning on the KAIST bearing dataset, the model achieved a test set accuracy of 98\%, with precision, recall, and F1-score all reaching 98\%, as summarized in Table~\ref{tab:kaist_perf}.

\begin{table}[h!]
\centering
\caption{Performance of the proposed model on the KAIST dataset.}
\label{tab:kaist_perf}
\begin{tabular}{lcccc}
\hline
\textbf{Dataset} & \textbf{Accuracy} & \textbf{Precision} & \textbf{Recall} & \textbf{F1-Score} \\
\hline
KAIST & 0.98 & 0.98 & 0.98 & 0.98 \\
\hline
\end{tabular}
\end{table}

\subsection{PINN Transfer Learning Strategies}
To investigate the effectiveness of integrating domain knowledge into TL models for bearing fault classification, a comprehensive analysis of physics-informed TL models under three different operating conditions is presented in Tables~\ref{tab:physinf_oc1}--\ref{tab:physinf_oc3}. 
In Operating Condition~1 (N09\_M07\_F10), both approaches attained the best results using Model~2, where the non-physics-informed model achieved slightly less accuracy (0.92~$\pm$~0.01 compared to 0.91~$\pm$~0.01) than the physics-informed model. In contrast, for the other two models, physics-informed TL models achieved an accuracy of 0.88~$\pm$~0.01, surpassing their non-physics-informed alternatives (0.81~$\pm$~0.01 for both models).

For Operating Condition~2 (N15\_M01\_F10), the physics-informed Model~2 reached the highest scores (0.97~$\pm$~0.01 for all metrics), slightly outperforming the non-physics-informed Model~2 (0.96~$\pm$~0.01). Additionally, Model~1 and Model~3 in the physics-informed context exhibited enhanced performance (0.96~$\pm$~0.01 and 0.95~$\pm$~0.01, respectively), in contrast to the non-physics-informed counterparts (0.92~$\pm$~0.01 and 0.81~$\pm$~0.01, respectively). 

The physics-informed Model~2 maintained its superiority over its non-physics-informed version in Operating Condition~3 (N15\_M07\_F04), achieving an accuracy of 0.93~$\pm$~0.01 versus 0.89~$\pm$~0.01. In comparison to the non-physics-informed approach (0.83~$\pm$~0.01 for both), Model~1 and Model~3 in the physics-informed approach also maintained a more stable and higher performance (0.90~$\pm$~0.01 for both).

In a nutshell, these findings demonstrate that the incorporation of physics-informed features in TL models results in substantial improvements in accuracy, precision, recall, and F1 score, along with enhanced robustness across different operating conditions and TL strategies. The models' improved and more consistent performance can be attributed to the integration of domain knowledge, which allows them to more accurately capture the fundamental physical mechanisms of bearing faults. Furthermore, the significance of integrating data-driven and physics-informed strategies for reliable and effective bearing fault classification in real-world applications is emphasized by the superior generalization capability and robustness to changes in operating environments that physics-informed TL models exhibit.

\begin{table}[h!]
\centering
\caption{Performance of three physics-informed TL Models with a confidence interval $\alpha = 0.05$ on N09\_M07\_F10 (Operating Condition 1).}
\label{tab:physinf_oc1}
\begin{tabular}{lcccc}
\hline
\textbf{Model} & \textbf{Accuracy} & \textbf{Precision} & \textbf{Recall} & \textbf{F1 Score} \\
\hline
TSFT (Model 1) & 0.88 $\pm$ 0.01 & 0.88 $\pm$ 0.01 & 0.88 $\pm$ 0.01 & 0.88 $\pm$ 0.01 \\
LAS (Model 2)  & 0.91 $\pm$ 0.01 & 0.91 $\pm$ 0.01 & 0.91 $\pm$ 0.01 & 0.91 $\pm$ 0.01 \\
HFR (Model 3)  & 0.88 $\pm$ 0.01 & 0.88 $\pm$ 0.01 & 0.88 $\pm$ 0.01 & 0.88 $\pm$ 0.01 \\
\hline
\end{tabular}
\end{table}

\begin{table}[h!]
\centering
\caption{Performance of three physics-informed TL Models with a confidence interval $\alpha = 0.05$ on N15\_M01\_F10 (Operating Condition 2).}
\label{tab:physinf_oc2}
\begin{tabular}{lcccc}
\hline
\textbf{Model} & \textbf{Accuracy} & \textbf{Precision} & \textbf{Recall} & \textbf{F1 Score} \\
\hline
TSFT (Model 1) & 0.96 $\pm$ 0.01 & 0.96 $\pm$ 0.01 & 0.96 $\pm$ 0.01 & 0.96 $\pm$ 0.01 \\
LAS (Model 2)  & 0.97 $\pm$ 0.01 & 0.97 $\pm$ 0.01 & 0.97 $\pm$ 0.01 & 0.97 $\pm$ 0.01 \\
HFR (Model 3)  & 0.95 $\pm$ 0.01 & 0.95 $\pm$ 0.01 & 0.95 $\pm$ 0.01 & 0.95 $\pm$ 0.01 \\
\hline
\end{tabular}
\end{table}

\begin{table}[h!]
\centering
\caption{Performance of three physics-informed TL Models with a confidence interval $\alpha = 0.05$ on N15\_M07\_F04 (Operating Condition 3).}
\label{tab:physinf_oc3}
\begin{tabular}{lcccc}
\hline
\textbf{Model} & \textbf{Accuracy} & \textbf{Precision} & \textbf{Recall} & \textbf{F1 Score} \\
\hline
TSFT (Model 1) & 0.90 $\pm$ 0.01 & 0.90 $\pm$ 0.01 & 0.90 $\pm$ 0.01 & 0.90 $\pm$ 0.01 \\
LAS (Model 2)  & 0.93 $\pm$ 0.01 & 0.93 $\pm$ 0.01 & 0.93 $\pm$ 0.01 & 0.93 $\pm$ 0.01 \\
HFR (Model 3)  & 0.83 $\pm$ 0.01 & 0.83 $\pm$ 0.01 & 0.83 $\pm$ 0.01 & 0.83 $\pm$ 0.01 \\
\hline
\end{tabular}
\end{table}
\subsection{Hypothesis Evaluation}
\label{subsec:hypothesis_evaluation}

To evaluate the hypothesis that transfer learning with fine-tuning on target operating conditions significantly improves model performance on unseen operating conditions, independent t-tests were performed comparing the classification accuracies of the transfer learning model (Model~2) and the baseline model across three different operating conditions. Table~\ref{tab:hypothesis_results} shows that the physics-informed transfer learning model (Model~2) achieved significantly higher accuracy, with a $p$-value less than 0.01 for all instances. These results suggest that physics-informed transfer learning by fine-tuning on new unseen operating conditions substantially improves the model’s ability to generalize, thereby leading to more robust fault classification across diverse operating conditions.

\begin{table}[h!]
\centering
\caption{Statistical comparison of baseline and transfer learning (Model~2) classification accuracies when tested on unseen operating conditions using independent t-tests ($n = 10$).}
\label{tab:hypothesis_results}
\begin{tabular}{lccc}
\hline
\textbf{Comparison} & \textbf{t-statistic} & \textbf{$p$-value} & \textbf{Interpretation} \\
\hline
N09\_M07\_F10: Baseline vs Model~2 & 160.95 & $<$ 0.01 & Statistically significant \\
N15\_M01\_F10: Baseline vs Model~2 & 15.79  & $<$ 0.01 & Statistically significant \\
N15\_M07\_F04: Baseline vs Model~2 & 45.60  & $<$ 0.01 & Statistically significant \\
\hline
\end{tabular}
\end{table}

\section{Discussion}
\label{sec:discussion}

The experimental results demonstrate that the proposed multimodal 1D CNN framework, integrating vibration and motor current signals, achieves high performance in bearing fault classification under varying operating conditions. The baseline model, trained solely on the baseline condition, attained accuracies of $0.40 \pm 0.01$, $0.90 \pm 0.01$, and $0.76 \pm 0.01$ for operating conditions N09\_M07\_F10, N15\_M01\_F10, and N15\_M07\_F04, respectively (Table~\ref{tab:baseline_vs_conditions}). Although precision, recall, and F1 scores followed a similar trend, the relatively low performance in N09\_M07\_F10 highlighted the challenges of domain shift when training and testing conditions differ significantly.

The application of transfer learning (TL) markedly improved generalization across unseen conditions. Among the tested strategies, Model~2 (freezing convolutional and pooling layers while fine-tuning the fully connected layer) achieved the best results, with accuracy improvements up to 0.08 compared to the baseline. For instance, in N09\_M07\_F10, TL using Model~2 achieved $0.92 \pm 0.01$ accuracy, outperforming the baseline by over $50\%$ relative improvement. This confirms that preserving low-level feature extraction while adapting higher-level decision layers is an effective approach for cross-domain fault diagnosis.

The t-SNE visualizations (Figures~\ref{fig:tsne_baseline} and \ref{fig:tsne_tl}) provided further insight into the internal feature representations. For the baseline model, feature clusters corresponding to Healthy, Inner Fault, and Outer Fault categories showed partial overlap, indicating less distinct separation. After TL adaptation, the clusters became more compact and well-separated, suggesting that the adapted models learned more discriminative and domain-invariant features.

Validation on the KAIST dataset demonstrated strong cross-dataset generalization. Using Model~1 TL strategy, the proposed model achieved an accuracy of $0.98$, with matching precision, recall, and F1 scores ($0.98$ each) despite differences in operating speed (3010~rpm), load torque (0~Nm), and radial force (0~N) compared to the Paderborn dataset. This result confirms that the architecture is transferable across different experimental setups without significant performance degradation.

The incorporation of physics-informed features into the TL models consistently improved robustness and accuracy across all operating conditions (Tables~\ref{tab:physinf_oc1}--\ref{tab:physinf_oc3}). For example, in N15\_M01\_F10, the physics-informed Model~2 achieved $0.97 \pm 0.01$ accuracy, compared to $0.96 \pm 0.01$ for the non-physics-informed counterpart. In N09\_M07\_F10, physics-informed Model~1 and Model~3 both achieved $0.88 \pm 0.01$, surpassing the $0.81 \pm 0.01$ performance of their non-physics-informed equivalents. These consistent gains highlight the advantage of embedding domain knowledge into data-driven models, enabling them to capture the underlying physical mechanisms of bearing faults more effectively.

Importantly, the improvements observed with transfer learning and physics-informed modeling were not only consistent but also statistically significant. Independent $t$-tests comparing the physics-informed TL model (Model~2) with the baseline across all three unseen operating conditions yielded $p < 0.01$ in every case (Table~\ref{tab:hypothesis_results}), confirming that the observed gains are unlikely to be due to random variation. This statistical validation reinforces the robustness of the proposed framework for generalizable and interpretable bearing fault diagnosis in real-world industrial environments.

\subsection{Limitations and Future Research Directions}
While the proposed multimodal 1D CNN framework demonstrated strong generalization across operating conditions and datasets, several limitations remain.
First, the experiments were conducted on controlled laboratory datasets (Paderborn University and KAIST), which may not fully capture the noise, sensor degradation, and environmental variability present in real-world industrial environments.
Second, although transfer learning substantially improved performance (e.g., from $0.40 \pm 0.01$ to $0.92 \pm 0.01$ accuracy in N09\_M07\_F10), the adaptation process still required labeled data from the target domain, which may be scarce in practice.
Third, the physics-informed models enhanced robustness by up to $0.07$ in accuracy compared to their non-physics-informed counterparts, but their effectiveness depends on the accuracy and completeness of the incorporated physical knowledge, which might not always be readily available.

Future research could address these limitations by:
\begin{itemize}
    \item Developing unsupervised or few-shot domain adaptation methods to minimize dependence on labeled target data.
    \item Validating the framework on large-scale, real-time industrial monitoring systems with diverse machinery types.
    \item Incorporating additional modalities such as temperature, acoustic emission, or oil debris analysis.
    \item Integrating adaptive physics-informed constraints that can dynamically adjust to evolving operating conditions.
    \item Exploring online learning strategies to update the model continuously as new data streams in.
\end{itemize}
\section{Conclusion}
This study introduced a multimodal 1D CNN framework for bearing fault classification that combines vibration and two-phase motor current signals, integrates physics-informed features, and leverages multiple transfer learning (TL) strategies to improve adaptability under variable operating conditions. Using the Paderborn University and KAIST datasets, the proposed approach demonstrated substantial gains over baseline single-modality and non-physics-informed methods. In the most challenging unseen condition (N09\_M07\_F10), TL improved accuracy from $0.40 \pm 0.01$ to $0.92 \pm 0.01$, representing a relative increase of more than $50\%$. Across all tested operating conditions, the incorporation of physics-informed features further enhanced robustness, with improvements of up to $0.07$ in accuracy compared to non-physics-informed counterparts. For instance, in N15\_M01\_F10, the physics-informed Model~2 achieved $0.97 \pm 0.01$ accuracy versus $0.96 \pm 0.01$ for its non-physics-informed equivalent. Hypothesis testing using independent t-tests confirmed that these improvements were statistically significant ($p < 0.01$) across all unseen operating conditions. Cross-dataset evaluation, where models trained on Paderborn data were tested on KAIST, yielded accuracies exceeding $0.90$ in two out of three operating conditions, confirming strong generalization capability. These findings indicate that fusing complementary sensing modalities, selectively fine-tuning pretrained models, and embedding domain-specific physics not only improve classification performance but also enhance model stability under severe domain shifts. The proposed framework shows high potential for deployment in real-world industrial condition monitoring systems, where operating conditions are dynamic and labeled target-domain data are often scarce.

\section*{Conflict of interest}
The authors declare no conflict of interest.
\bibliographystyle{unsrt}  
\bibliography{main} 
\appendix
\section{Appendix}
\section{Transfer Learning Strategies}
\label{appendix:tl_strategies}

\subsection*{Layer-Wise Adaptation Strategy (LAS)}

In LAS (Model~2), the first convolution and pooling layers are frozen, while all subsequent layers, including the fully connected layer, are fine-tuned on the target domain~\cite{liu2021domain,wen2020transfer}. This approach retains low-level domain-invariant features while adapting higher-level task-specific representations to new operating conditions.

The model parameters can be expressed as:
\[
\theta = (\theta_f, \theta_m, \theta_c)
\]
where:
\begin{itemize}
    \item $\theta_f$: frozen base layers
    \item $\theta_m$: partially trainable middle layers
    \item $\theta_c$: fully trainable classification layers
\end{itemize}

The optimization objective is:
\[
(\theta_m^*, \theta_c^*) = \arg\min_{\theta_m, \theta_c} \frac{1}{N} \sum_{i=1}^{N} L\big(f(x_i; \theta_f, \theta_m, \theta_c), y_i\big)
\]
where $N$ is the number of training samples in the target domain.

\begin{figure}[h]
    \centering
    \includegraphics[width=0.8\textwidth]{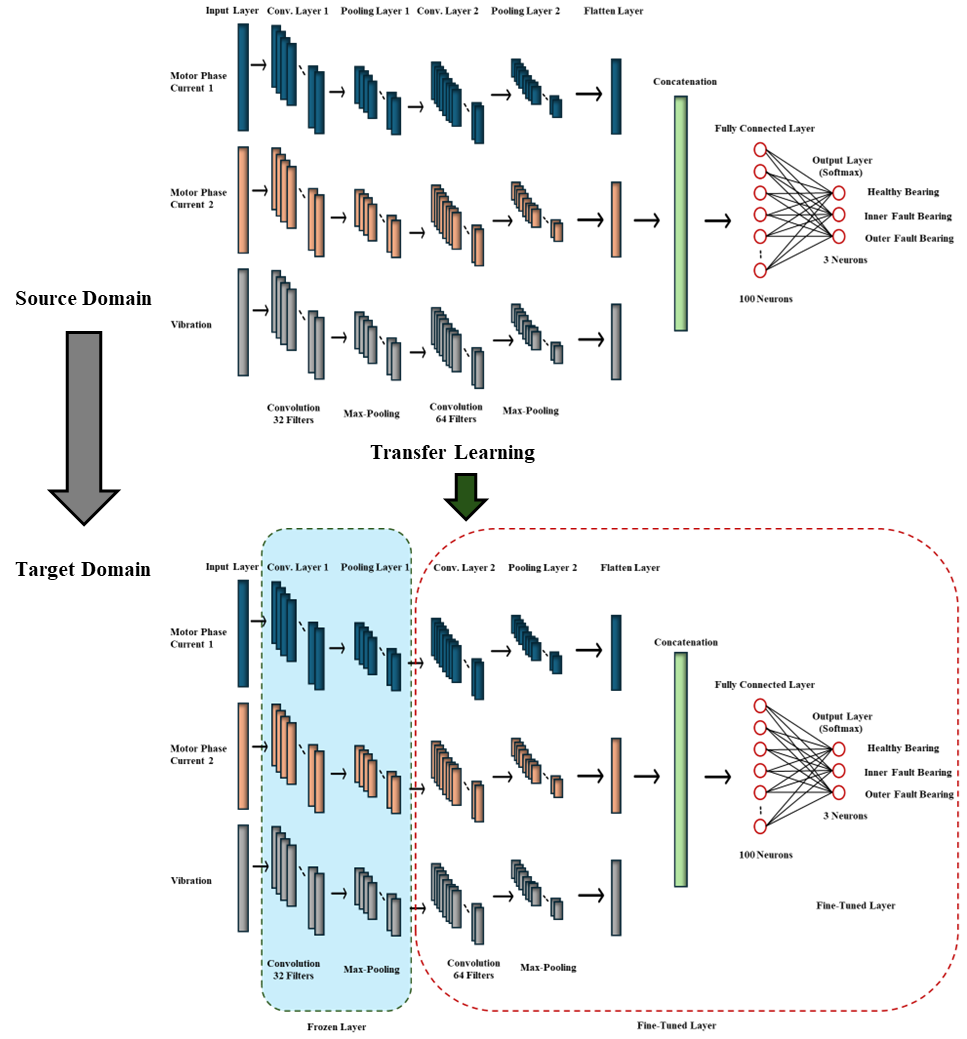}
    \caption{Framework of the proposed TL Model~2 (Layer-Wise Adaptation Strategy).}
    \label{fig:las}
\end{figure}

\subsection*{Hybrid Feature Reuse (HFR)}

In HFR (Model~3), all layers before the fully connected stage are frozen, and the final classifier is replaced with a new target-specific layer~\cite{wen2020transfer}. This ensures that previously learned features are retained while enabling adaptation to the target domain via the newly introduced classifier.

The model parameters can be expressed as:
\[
\theta = (\theta_r, \theta_t)
\]
where:
\begin{itemize}
    \item $\theta_r$: frozen reused feature extractor layers
    \item $\theta_t$: new trainable classification layers
\end{itemize}

The optimization objective is:
\[
\theta_t^* = \arg\min_{\theta_t} \frac{1}{N} \sum_{i=1}^{N} L\big(f(x_i; \theta_r, \theta_t), y_i\big)
\]

\begin{figure}[h]
    \centering
    \includegraphics[width=0.8\textwidth]{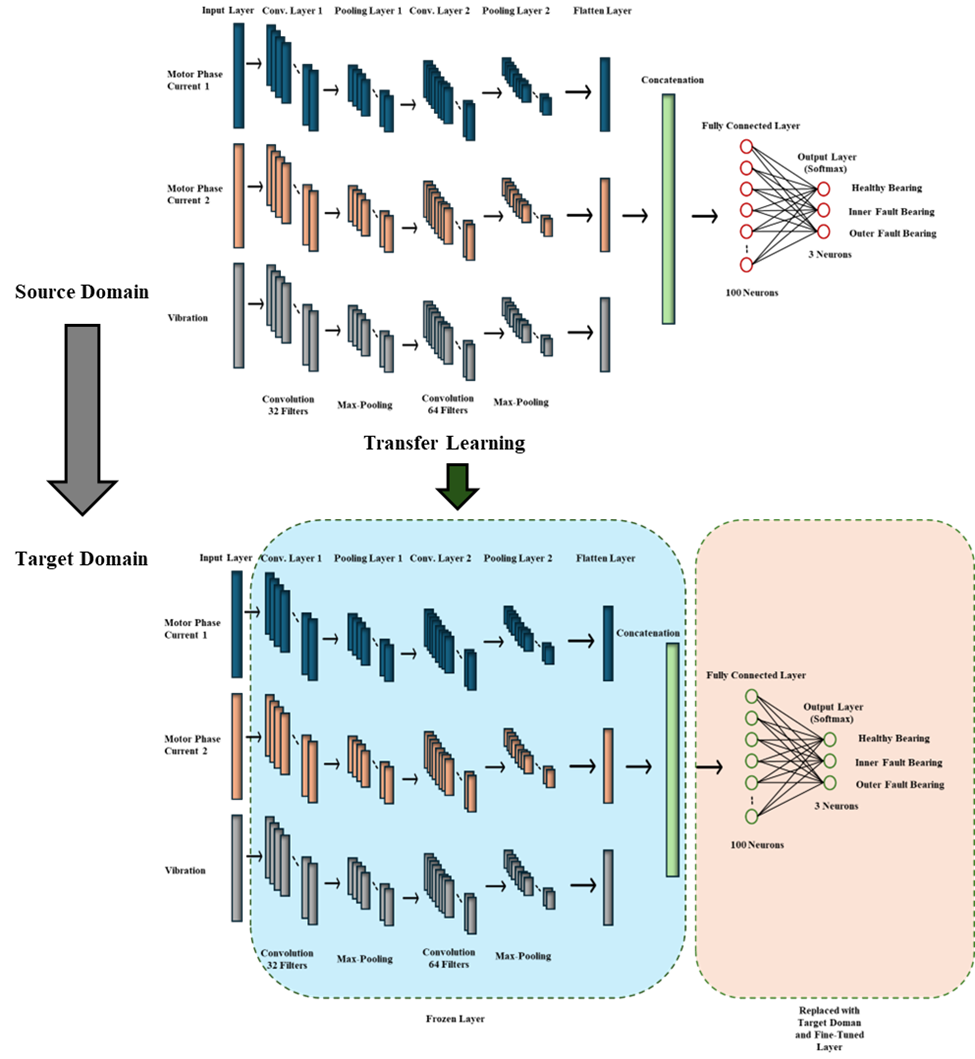}
    \caption{Framework of the proposed TL Model~3 (Hybrid Feature Reuse).}
    \label{fig:hfr}
\end{figure}

\end{document}